\documentclass[twocolumn, times, trackchanges]{aastex63_bren}

\usepackage[ruled,vlined]{algorithm2e}
\usepackage{graphicx}
\usepackage{bbm}
\usepackage{hyperref}
\usepackage{subfiles} 

\begin{document}

\title{Extracting the main trend in a dataset: the Sequencer algorithm}
\author{Dalya Baron}
\affiliation{School of Physics and Astronomy, Tel-Aviv University, Tel Aviv 69978, Israel}
\author{Brice M\'enard}
\affiliation{Department of Physics and Astronomy, The Johns Hopkins University, 3400 N Charles Street, Baltimore, MD 21218, USA}
\date{\today}
\email{dalyabaron@gmail.com, menard@jhu.edu}

\begin{abstract}
Scientists aim to extract simplicity from observations of the complex world. An important component of this process is the exploration of data in search of trends. In practice, however, this tends to be more of an art than a science. Among all trends existing in the natural world, one-dimensional trends, often called sequences, are of particular interest as they provide insights into simple phenomena. However, some are challenging to detect as they may be expressed in complex manners. We present the Sequencer, an algorithm designed to generically identify the main trend in a dataset. It does so by constructing graphs describing the similarities between pairs of observations, computed with a set of metrics and scales. Using the fact that continuous trends lead to more elongated graphs, the algorithm can identify which aspects of the data are relevant in establishing a global sequence. Such an approach can be used beyond the proposed algorithm and can optimize the parameters of any dimensionality reduction technique. We demonstrate the power of the Sequencer using real-world data from astronomy, geology as well as images from the natural world. We show that, in a number of cases, it outperforms the popular t-SNE and UMAP dimensionality reduction techniques. This approach to exploratory data analysis, which does not rely on training nor tuning of any parameter, has the potential to enable discoveries in a wide range of scientific domains. The source code is available on github and we provide an online interface at \url{http://sequencer.org}.
\end{abstract}


\section{Introduction}

``One of the principal objects of theoretical research is to find the point of view from which the subject appears in the greatest simplicity", wrote Josiah Willard Gibbs in 1881. The early phase of this process often involves exploratory data analysis, i.e. a search for patterns in a dataset without the benefit of guidance from theory. Unfortunately, data can appear complex and might not allow underlying trends to be revealed straightforwardly. Additional challenges include high dimensionality, the presence of noise, and ever-growing data volumes, all of which prevent efficient visualization of the data and require mathematically guided exploration. 

Dimensionality reduction techniques, from Principal Component Analysis (PCA; \citealt{pearson1901}) to the more recent t-Distributed Stochastic Neighbor Embedding (t-SNE; \citealt{vanDerMaaten08}) and Uniform Manifold Approximation and Projection for Dimension Reduction (UMAP; \citealt{mcInnes18}), provide powerful ways to address some of these limitations \citep{vanDerMaaten09, lee10, venna10}. However, scientific work does not end once dimensionality reduction algorithms have been applied to a dataset. Rather, it only begins. Extracting simplicity through the identification of interesting trends critically relies on user input and judgement. Domain knowledge informs important decisions: the choice of coordinates, the scale to focus on, the metric to use to compare objects, etc. In addition, when the size or dimensionality of data objects is large, such is the case for images, spectra or time series, analyses are typically performed on extracted features or summary statistics rather than on measured values or pixels. This decision is critical in the exploration process since a poor choice of summary statistics or relevant features might prevent the detection of interesting trends in the data. Finally, most dimensionality reduction algorithms depend on parameters, and changing them can dramatically affect the resulting representation 
\citep{wattenberg16, mcInnes18, baron19b}. 
Thus, a crucial part of the scientific exploration process is devoted to understanding which observables or summary statistics to investigate, which method to pick, and which parameter values to use, in order to obtain the most interesting results. In many cases, we lack a well-defined metric to guide these decisions (e.g., \citealt{lee10, zhang11, baron19b}). Without theoretical guidance, trial and error is usually the adopted strategy. Is it possible to perform these tasks in a way that will automatically and robustly identify the existence of a simple trend in an apparently complex dataset? Is it possible to operate directly on the raw data, without specifying which observables or summary statistics to use? 

When trying to understand a dataset, one attempts to extract meaning by building a generic and concise representation of it. For the representation to be generic, it should be invariant to coordinates transformation and deformations. Meaningful trends are topological properties of the data, i.e. aspects of the data manifold that are not affected by deformation \citep{carlsson2009topology}. For example, clusters can often be defined and interpreted, irrespective of the choice of coordinates. Many cluster finding algorithms are available and widely used (e.g., \citealt{ward63, macqueen67, cheng95, ester96, rodriguez14}). However, in many cases, we expect observed phenomena to exhibit a continuous change in their properties as a function of a leading, possibly unknown, driving parameter. In other words, we often expect to find sequences -- they abound in the natural world and, especially, in scientific measurements. Although they are essentially one-dimensional trends, they are often challenging to find as they can be expressed in complex manners.

In this paper we present an algorithm to automatically detect sequences in datasets. It uses information about the shape of the graph describing the similarities between the objects and exploits the fact that sequences give rise to elongated graphs. Following this approach, it is possible to consider different representations of the data and, for each of them, quantify the degree to which a continuous trend exists. We show that this method can also be used to optimize some of the parameters or certain arbitrary choices involved in dimensionality reduction techniques aimed at detecting sequences. Importantly, this search can be performed directly on the pixels or measured values, as opposed to user-defined observables or restrictive summary statistics. Therefore, our approach enables a more generic  search for continuous trends in arbitrary datasets.

\section{The signature of a continuous trend}
\label{s:signature_of_a_sequence}

\begin{figure*}
\begin{centering}
\includegraphics[width=1\linewidth]{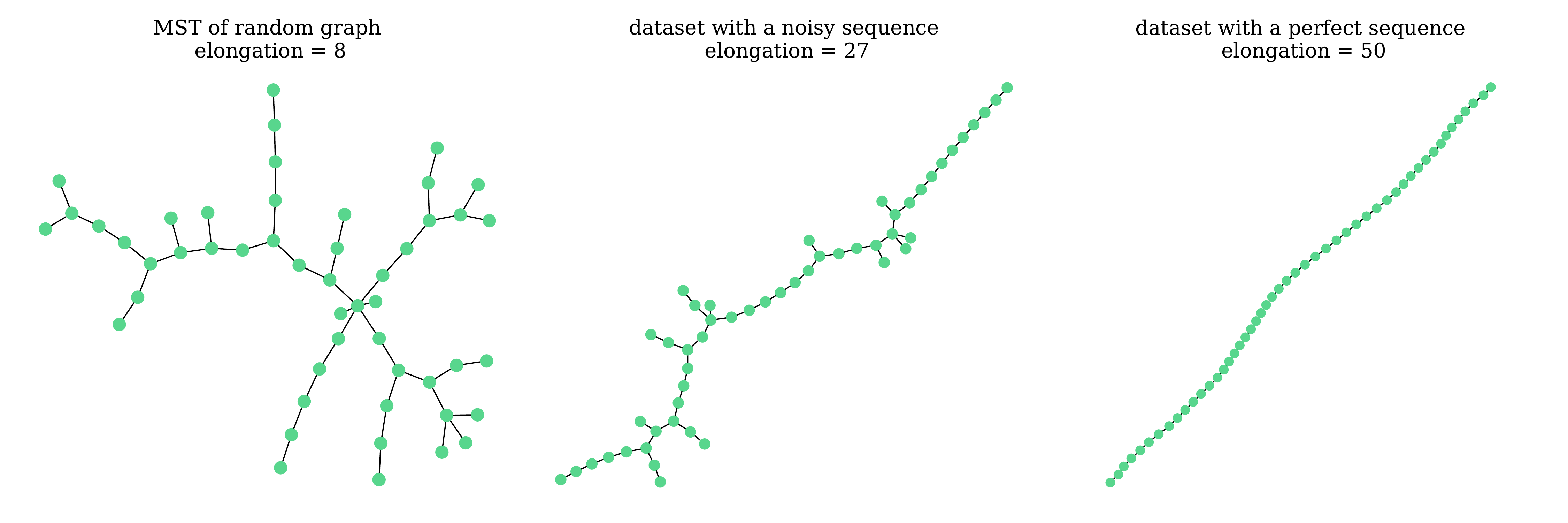}
\caption{Minimum spanning trees for three sets of 50 objects for which the similarities range from random (left) to continuous (right). Each panel shows the corresponding elongation parameter $\eta$ (Eq.~\ref{eq:elongation}) which can be used to characterize the degree to which a continuous trend is detected.} 
\label{f:shape_of_MST}
\end{centering}
\end{figure*}

Data acquisition often provides us with a collection of objects that are not necessarily ordered in a meaningful manner. If these objects follow an underlying trend due to the variation of an intrinsic parameter, it should be possible to order the set meaningfully. If the variation of this parameter leads to a continuous change of observables, the ordered set should minimize the cumulative differences between consecutive objects. Finding the ordering leading to such minimization is therefore expected to reveal the leading trend in a dataset. By doing so, we face several challenges: which aspects of the data should be considered? Which pixels carry relevant information? How to meaningfully define similarities between them? 

To address these questions, we propose the following approach: let us consider a collection of N objects and let us first assume that we have a useful metric allowing us to estimate the similarity (or equivalently a distance) between each pair of objects. The corresponding adjacency matrix represents a fully-connected graph, where each object in the original dataset is represented by a node in the graph, and the weights of the edges that connect the nodes represent the distances between the objects. This structure encodes all the possible trajectories within the dataset.

Within this set, some trajectories are of special interest: those that connect all the nodes and minimize the total distance accumulated along them\footnote{This is not equivalent to solving the Traveling Salesman's Problem. The Traveling Salesman's Problem is more specific and aims at finding a trajectory which starts and ends at the same node and visits all the others only once.}. We can find such a trajectory by finding the minimum spanning tree (MST) of our graph, the subset of the edges in a fully-connected graph that connects all the nodes together, without any cycles, and with the minimum possible \emph{total} edge weight (e.g., \citealt{kruskal56}). If each edge in the original fully-connected graph has a distinct weight, then the minimum spanning tree is unique. The shape of the resulting tree carries valuable information on the topological properties of the dataset. In particular, it can be used as an indicator of the degree to which a sequence exists in the dataset. In Figure \ref{f:shape_of_MST} we show examples of minimum spanning trees for three scenarios. The left panel shows the minimum spanning tree of a random graph. The middle panel shows the minimum spanning tree of a dataset with a noisy sequence, and the right panel shows the minimum spanning tree of a dataset with a perfect sequence. The existence of a continuous trend leads to a more elongated minimum spanning tree. Therefore, \emph{the minimum spanning tree elongation can characterize the existence of a trend} underlying a collection of objects.

When doing exploratory data analysis, one typically does not know a-priori how to meaningfully ``look'' at the data. Which similarity measure will be informative? On which scales will we find relevant information? Interestingly, we can address these questions by simply considering, in each case, the shape of the corresponding minimum spanning tree. In other words, we can automatically find the parameters that are most sensitive to the existence of a simple trend in the dataset. For a diverse enough set of distance metrics and scales, this automatic process has the capacity to reveal trends in a generic manner.
These trends can be intrinsic to the objects in the sample and/or extrinsic and driven by observational effects.

\section{Algorithm description}\label{s:algorithm_description}

We now provide a description of the Sequencer algorithm following the key principles outlined in the previous section. Our goal is to order a collection of $N_{\rm obj}$ objects with $N_{\rm pix}$ values. We will describe each object as $X^j_i$, with $j=1$ to $N_{\rm obj}$ and $i=1$ to $N_{\rm pix}$. To characterize and extract a sequence underlying such a collection of objects, we proceed in two steps:
\begin{enumerate}
    \item For a list of metrics and scales, we compute the corresponding graphs describing the dataset and quantify the elongation of their minimum spanning trees;
    \item We aggregate the results to form a new graph summarizing the relevant information and extract an ordered list of objects from it. 
\end{enumerate}
As we will demonstrate, this can reveal the trend characterizing the leading variation among the objects in many cases. Such a trend is often meaningful.

\subsection{Distance metrics \& scales}\label{s:algo:distance_metrics}

As described above, we can achieve a meaningful ordering of the dataset by minimizing the similarity or distance between adjacent objects. Doing so requires the choice of a distance measure. In order to be generic, we include several commonly-used metrics. This will allow the algorithm to consider various aspects of the data and its features. Similarly, we consider a list of scales on which the relevant information can be distributed.

By default, we use the following metrics: (i) the Euclidean Distance, (ii) the Kullback-Leibler Divergence (KL Divergence; \citealt{kullback51}), (iii) the Monge-Wasserstein or Earth Mover Distance (EMD; \citealt{rubner98}), and (iv) the Energy Distance \citep{szekely02}. The definitions and properties of these metrics are described in the Appendix. 
If desired by the user, this list can be expanded to better suit a particular application. We note that this default set includes metrics with different properties: the Earth Mover Distance and Energy Distance are sensitive to the magnitude of displacements along the $i$ coordinate. This is important with continuous measurements, for example performed as a function of space or time, for which the derivative with respect to $i$ carries relevant information. In contrast, the Euclidian Distance and the KL Divergence treat the different pixels of $X_i$ as different dimensions and are insensitive to index shuffling. They provide a qualitatively different view of the information content. 

The observable signature of an underlying trend can exist on different scales which may not be known a-priori. In order to be generic, it is important to consider a range of scales.  To do so, we decompose each object $X$ into a series of contiguous segments whose length is given by $N_{\rm pix} / 2^{l}$. This results in an ensemble of segments which allows us to look at each data object hierarchically, starting from its entirety ($l=0$) and creating a binary tree such that the deepest scale corresponds to about twenty pixels. 
Thus, for a given metric $k$ and scale $l$, the object is split into $2^{l}$ segments, and we refer to each segment using the index $m$. The maximum depth or the ways in which the data is decomposed can be modified by the user if need be. 

\begin{algorithm}[t]
\SetAlgoLined
set list of metrics\;
set list of scales\;
\For{each metric $k$}{
    \For{each scale $l$}{
        set list of segments\;
        split objects $X$ into $m$ segments $X_{m}$\;
        normalize each segment to have a sum of 1\;
        \For{each segment $m$}{
            $D_{klm}$ = distance matrix($X_{m}$)\;
            $\mathrm{MST}_{klm}$ = Minimum Spanning Tree($D_{klm}$)\;
            $\eta_{klm} = a_{klm}/b_{klm} = {\rm elongation}(\mathrm{MST}_{klm})$\;
            }
        $D_{kl}$ = $\eta$-weighted average of individual $D_{klm}$\;
        $\mathrm{MST}_{kl}$ = Minimum Spanning Tree($D_{kl}$)\;
        $\eta_{kl}$ = elongation(MST$_{kl}$)\;
    }
}
$P =$ combined proximity matrix populated by $\eta$-weighted edges of $\mathrm{MST}_{kl}$ \;
\For{each pair of objects $i,j$}{
    $D_{ij}^{\rm combined} = {1}/{P_{ij}^{\rm combined}}$;
}
$\mathrm{MST}$ = Minimum Spanning Tree($D$)\;
Sequence = Breadth First Search path($\mathrm{MST}$)\;
\caption{Sequencer pseudo-code}
\end{algorithm}

\subsection{Finding the main sequence}\label{s:algo:graph_properties}

Our goal is to look at the data for each metric $k$, scale $l$ and segment $m$, and estimate the level to which an underlying trend is present. To do so, we proceed as follows. First, for each scale $l$ and each segment $m$, we first normalize each object such that the sum over its components is one. We then extract geometrical properties of the set of graphs characterizing similarities between all pairs of objects:

\indent\textbf{$\bullet$ Graph minimum spanning tree:} 
for each metric $k$, scale $l$ and segment $m$, we compute a $N_{\rm obj}^2$ distance matrix $D_{klm}$ which represents a fully-connected graph. We then compute its minimum spanning tree using Kruskal's algorithm \citep{kruskal56}. It gives us a set of $k\times l \times m$ trees with $N_{\rm obj}$ nodes. The key information on the presence of an underlying trend resides in the shape of these graphs.

\textbf{$\bullet$ Graph length: } The least connected node, $j_{\rm LC}$, of the minimum spanning tree is expected to belong to its longest branch. To identify it, we compute the closeness centrality of each node \citep{freeman78} and select the one with the smallest value. We then compute the shortest path $\Delta_{klm}(j_{\rm LC},j)$ between this node and every other node in the minimum spanning tree. The shortest path is a unitless integer counting the minimal number of edges between the two nodes (see the Appendix for additional details). We then define the major axis of the minimum spanning tree to be the average of the shortest paths over all nodes:
\begin{equation}
\label{eq:graph_length}
a_{klm} = \langle \Delta_{klm}(j_{\rm LC},j) \rangle_{ {\rm node}\;j}\;.
\end{equation}

\textbf{$\bullet$ Graph width: } Each node in the graph can be assigned to a level $q$ which corresponds to a unique value of $\Delta_{klm}(j_{\rm LC},j)$. That is, the shortest path between all the nodes that are assigned to the level $q$ and the least connected node: $\Delta_{klm}(j_{\rm LC},j) = \Delta_{q}$
(see section~\ref{s:graph_nomenclature} for an illustration). The width of a level $q$, $\Delta_{klm}^\perp(q)$, is defined as the number of nodes that are assigned to it.
We use the average of this quantity as an estimate of the average half width, or minor axis $b$, of the minimum spanning tree:
\begin{equation}
\label{eq:graph_width}
b_{klm} = \frac{1}{2}\,\langle \Delta_{klm}^\perp(q) \rangle_{ {\rm level}\;q}
\end{equation}

\textbf{$\bullet$ Graph elongation: } for each metric $k$, scale $l$, and segment $m$, we then define the elongation of the minimum spanning tree as its average height divided by its average width: 
\begin{equation}
\label{eq:elongation}
    \eta_{klm} ={\rm elongation}(D_{klm}) = \frac{a_{klm}}{b_{klm} }\,.
\end{equation}
This quantity can then be used to characterize the level to which the signature of a continuous trend is apparent in a given segment of the objects, through a given metric and on a given scale. Importantly, by being a ratio of numbers of edges, this parameter can be defined irrespective of the metric used. It is a summary statistics describing geometrical properties of the minimum spanning tree and, as a result, topological properties of the data.

\begin{figure*}
\begin{centering}
\includegraphics[width=\linewidth]{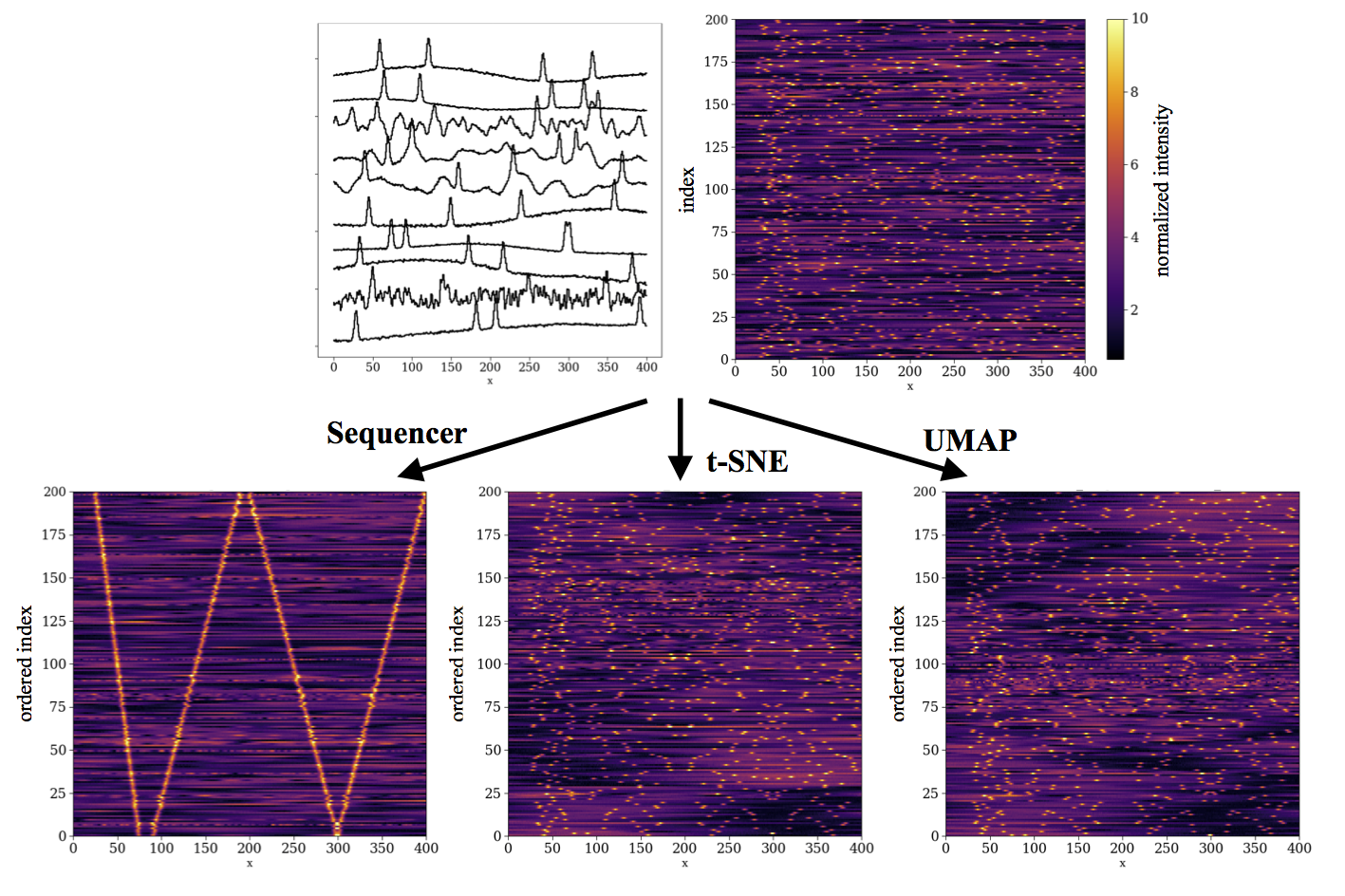}
\caption{\textbf{Application of the Sequencer, t-SNE, and UMAP to a simulated dataset with a clear one-dimensional sequence.} The top left panel shows examples of objects randomly selected from the simulated dataset. The top right panel shows all the objects from the dataset, where each row represents a single object and is color-coded according to the intensity in each of its pixels. The bottom panels show the objects, ordered according to the Sequencer, t-SNE, and UMAP respectively. It illustrates the ability of the Sequencer to focus on the scale of interest.}
\label{f:synthetic_dataset_example}
\end{centering}
\end{figure*}

\textbf{$\bullet$ Aggregation of scales and metrics:} each minimum spanning tree represents a sequence viewed through a given metric $k$ and scale $l$ of a segment of the data $m$. The elongation $\eta_{klm}$ of its corresponding minimum spanning tree carries information on the level at which an underlying trend is detected. We can first combine the information obtained for all segments by creating a global distance matrix $D_{kl}$ using an elongation-weighted average of our set of minimum spanning trees:
\begin{equation}
\label{eq:combine}
D_{kl} = \left\langle \eta_{klm}\,.\,D_{klm} \right\rangle_m\;.
\end{equation}
This provides us with $k\times l$ different ``views'' of the data, which we can attempt to aggregate. Here we need to keep in mind that different metrics $k$ will result in distance matrices $D_{kl}$ with different units. To meaningfully combine information obtained from different metrics, we will only extract the topological information of the resulting minimum spanning trees, as given by their edge counts, and use an elongation-weighted average to create a ``proximity'' matrix: 
\begin{equation}
P^{\rm combined} = 
\left\langle \eta_{kl}\,.\, \# {\rm ~of~edges}({\rm MST}(D_{kl})) \right\rangle_{kl}\;.
\label{eq:combine_metric}
\end{equation}
We set all the elements which are not populated by the edges of the minimum spanning trees to zero (no connection between the corresponding nodes), and the elements on the diagonal to infinity (the proximity of a node to itself is infinite). This then allows us to define a combined distance matrix whose elements $i,j$ are defined as $D_{ij}^{\rm combined} = {1}/{P_{ij}^{\rm combined}}$. This distance matrix $D^{\rm combined}$ provides us with a multi-scale and multi-metric characterization of the dataset. We then compute its minimum spanning tree and corresponding elongation. 

\indent\textbf{$\bullet$ Extraction of a final sequence:} in order to extract a sequence from the minimum spanning tree of the combined distance matrix, we must select a particular walk within the tree, i.e. we must select the relative order in which we visit all the nodes within the graph. As done above, we define the starting point of the sequence using the least connected node of the combined minimum spanning tree. From this starting point, we walk through the graph using a Breadth First Search (BFS; \citealt{cormen09}). If the combined minimum spanning tree presents an appreciable elongation, a BFS traversal is expected to define the main trend in the dataset. Thus, we expect the main branch of the tree to represent the sequence, and secondary branches to represent the scatter or the secondary sequence. 

We point out that the addition of one or several new objects to the dataset can be done without redoing the full process. Once a sequence has been obtained, one can easily insert a new object into it, by performing a straightforward neighbor search. More details are provided in the Appendix.

\subsection{Scaling considerations}
The algorithm described above requires distance matrix calculations which scale as $\mathcal{O}(N_{\rm obj}^2)$. For large datasets this approach can be computationally demanding. A useful alternative is to apply the technique to a subset of the data with $N_s\ll N$ objects, build the skeleton of a sequence and then populate it with the remaining points. This process leads to a faster computation which allows one to process significantly larger datasets, at the cost of obtaining only an approximate result. A more detailed description of this method is provided in the Appendix.

\begin{figure*}
\begin{centering}
\includegraphics[width=0.6\linewidth]{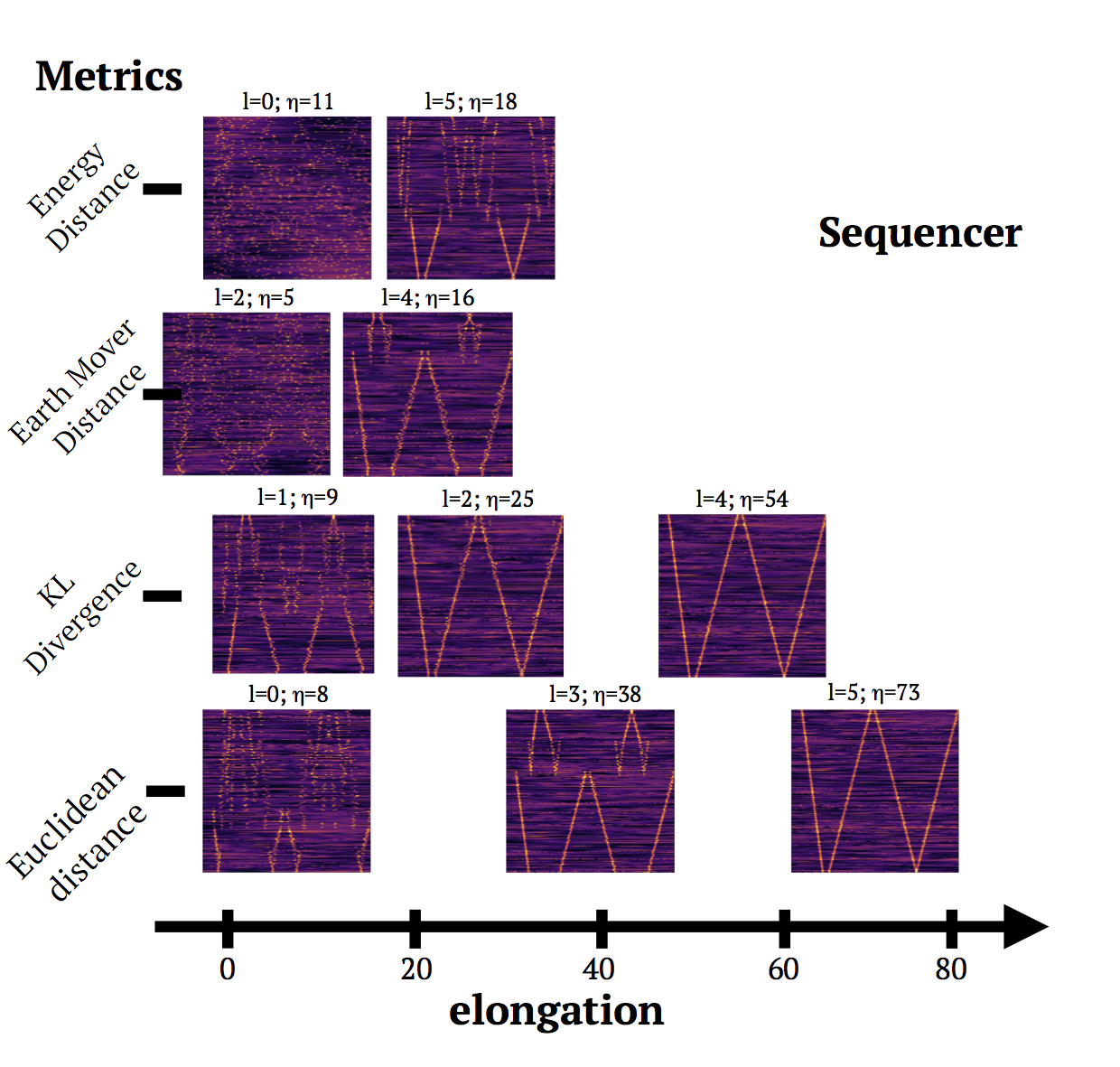}
\caption{\textbf{Using the elongation parameter to identify the scale and metric that reveal a meaningful trend in a dataset.} The synthetic dataset from figure \ref{f:synthetic_dataset_example} ordered according to individual distance metrics and scales. The scales and resulting elongation parameters are indicated at the top of each panel.}
\label{f:sequencer_scales_diagram_synthetic_dataset}
\end{centering}
\end{figure*}

\section{Performance \& results}\label{s:performance}

We now apply the algorithm to datasets of increasing complexity to demonstrate its effectiveness as well as its advantages compared to existing dimensionality reduction techniques. In this paper, for simplicity, we will only consider datasets with one dimensional objects. However, the key organizing principle based on the minimum spanning tree elongation can be applied to objects in two or higher dimensions\footnote{The publicly-available code can be applied to two-dimensional datasets.}, but at a higher computational cost. 
Unless specified otherwise, we apply the Sequencer using its default setting, i.e. with the four distance metrics and the binary scale decomposition described in section \ref{s:algorithm_description}. 

\begin{figure*}[t]
\begin{center}
\includegraphics[width=.95\linewidth]{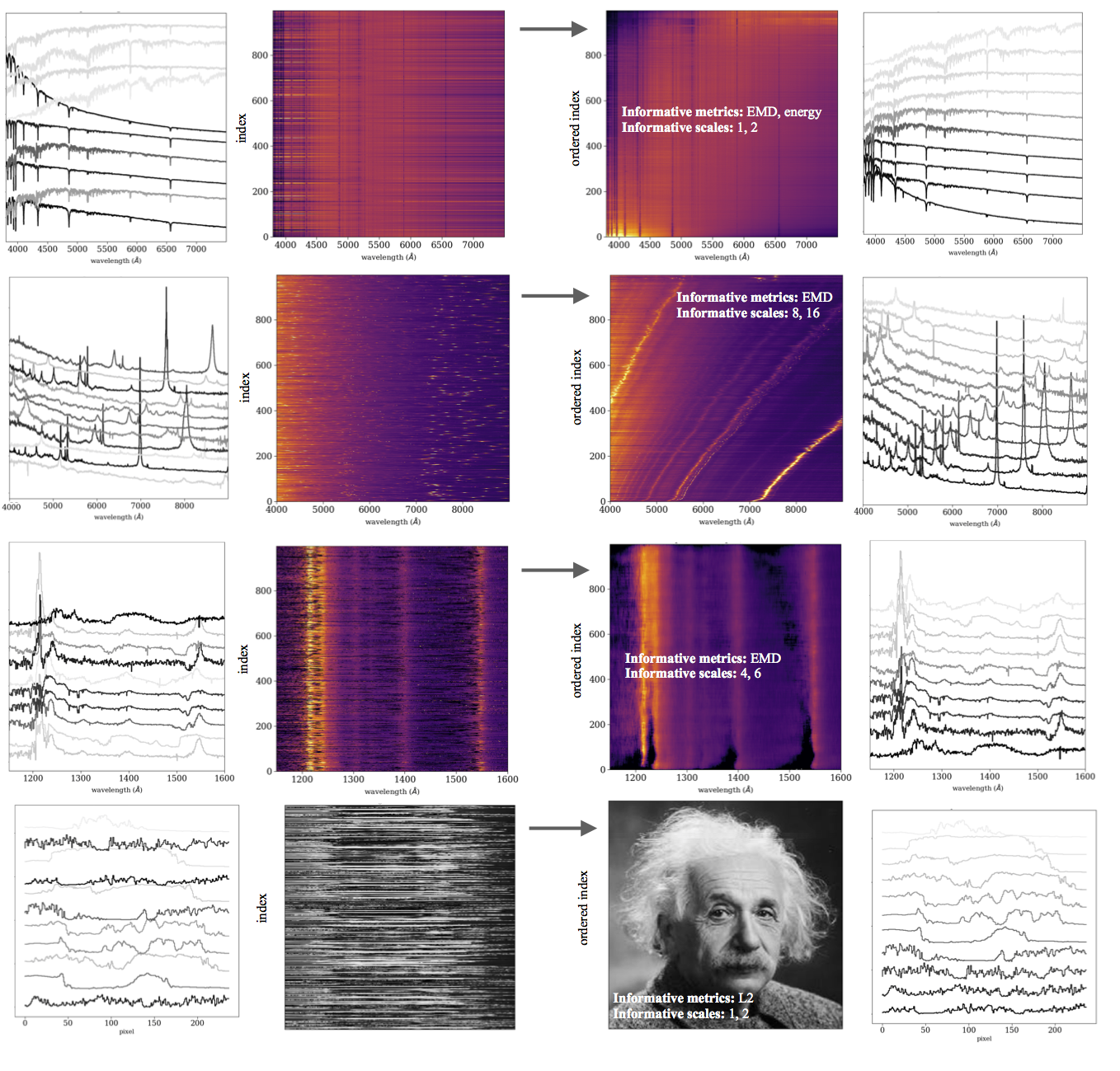}
\caption{
\textbf{Application of the Sequencer to four randomly-ordered datasets:}
(a) one thousand spectra of stars, which are then ordered by temperature,
(b) one thousand spectra of quasars, which are then ordered by redshift,
(c) one thousand spectra of quasars with complex broad absorption line systems, which are then ordered by absorber type and velocity distribution, and
(d) a picture of Albert Einstein with the rows shuffled, which is then properly reconstructed.
In each case, we indicate the metrics and the scales that resulted in the most elongated minimum spanning trees, and as a result, dominated the weighted averages in Eq~\ref{eq:combine} and \ref{eq:combine_metric}.
}
\label{f:sequences_physical_datasets}
\end{center}
\end{figure*}

\subsection{Validation with a simulated dataset}\label{s:performance:simulated_dataset}

We first construct a synthetic dataset with a well-defined trend that exists only on small scales, on top of a varying background. We create 200 one-dimensional objects with $N_{\rm pix}=400$ presenting four narrow Gaussian pulses whose positions vary continuously from one object to another to form a clear sequence. To this we add random large-scale fluctuations using a Gaussian process. We show the shuffled dataset in the top panels of Figure~\ref{f:synthetic_dataset_example}, where the left panel shows a subset of the objects in the sample, and the right panel visualizes the full dataset. Each row represents a different object, and the color-coding represents the relative intensity in each of its pixels. As can be seen, it is visually difficult to identify an underlying trend in this collection of objects. We apply the Sequencer to this dataset and show its output in the bottom left panel. The algorithm is capable of detecting the overall structure in the dataset even though only a small fraction of the pixels carries relevant information.

For comparison, we show how t-SNE and UMAP, two of the most popular dimensionality reduction techniques, perform on the same simulated dataset. To obtain a sequence using t-SNE or UMAP, we apply these techniques to embed the input dataset into one dimension, and then rank-order the objects according to their assigned value in this dimension. We point out that, as mentioned in the introduction, using these algorithms requires setting parameters. In practice, the user typically runs such an algorithm multiple times, varying those parameters until the ``best'' result is obtained. Here we present only the best sequences obtained after optimizing the distance metric in an automatic manner, using the elongation of the corresponding graphs. The details of this optimization are given in the Appendix. Clearly, t-SNE and UMAP fail to detect the sequence in narrow pulse locations as they only consider an object as a whole, without the ability to focus on individual segments and understand that a well-defined trend exists on small scales.

To illustrate how the Sequencer identifies the scale and metric revealing a meaningful trend, in Figure~\ref{f:sequencer_scales_diagram_synthetic_dataset} we present the resulting ordering of the algorithm for each metric and a set of scales as a function of the corresponding elongation parameter.
The most elongated minimum spanning trees are obtained using the Euclidean Distance and KL-Divergence measured over small scales ($l=4$ and $l=5$). These scales and metrics dominate the weighted averages in Eq~\ref{eq:combine} and \ref{eq:combine_metric} and, as a result, define the final ordering of the data.

\subsection{Examples with real datasets}\label{s:performance:scientific_datasets}

\begin{figure*}
\begin{center}
\includegraphics[width=1.\textwidth]{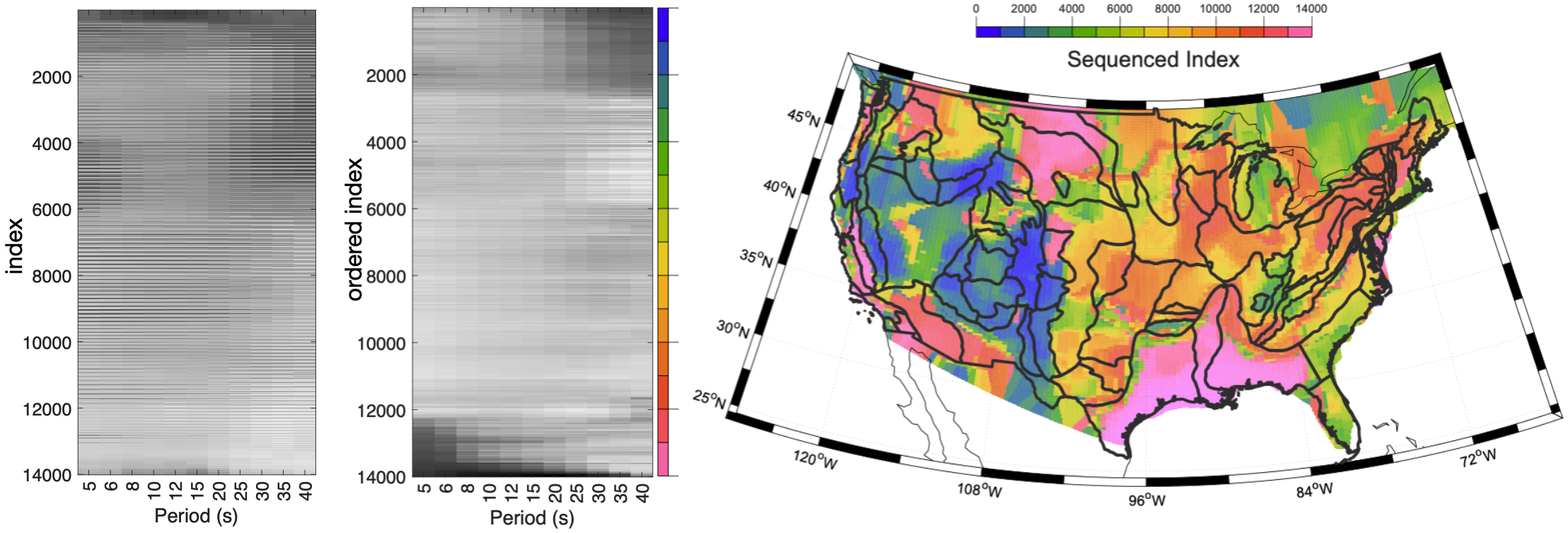}
\caption{\textbf{Application of the Sequencer to spatial mapping.} Left panels: normalized scores for Love wave phase velocities as a function of period extracted from maps by \citet{olugboji} at about $14,000$ locations beneath the contiguous United States and the same data after ordering by the Sequencer. Right panel: locations colored by their sequenced index, revealing geographic patterns which correspond well to physiographic provinces (black outlines, \citealt{physio}). Figure courtesy of Vedran Lekic.}
\label{f:US_map}
\end{center}
\end{figure*}

We now present a series of examples using one-dimensional data. For each dataset, we first display ten objects to convey the typical level of complexity. We then show a visualization of the entire randomly-ordered dataset, followed by the same data this time ordered by the Sequencer. We then display several objects from the ordered dataset. In each case, we indicate the metrics and scales that were identified as the most informative, based on their contributions to the minimum spanning tree elongation-weighted averages in Eq~\ref{eq:combine} and \ref{eq:combine_metric}.

\textbf{Spectroscopic data.} We start with a sample of $1,000$ spectra of stars from the publicly-available Sloan Digital Sky Survey (SDSS; \citealt{york00}). Each spectrum is a measurement of the brightness of a star as a function of wavelength. The dataset ordered by the Sequencer displays visible trends in both large-scale and small-scales features. A physical interpretation reveals that these continuous variations correspond to a sequence in the temperature of the stars. The top of the sequence is dominated by hot stars, which exhibit absorption lines due to hydrogen atoms, and the bottom part is dominated by cooler stars which exhibit absorption lines due to other, heavier, elements. 

We repeat the analysis with a set of $1,000$ spectra of quasars spanning a range of properties. This time we detect a trend in distance (redshift): the bright and narrow features shift continuously throughout the sequence. We point out that, using their default settings, most dimensionality reduction techniques (e.g., PCA, t-SNE, and UMAP) are insensitive to pixel shuffling and often fail to detect horizontal shifts in the data (see section~\ref{a:distances}).

Our third example shows another set of $1,000$ quasar spectra \citep{trump06}. The dark pixels correspond to flux deficits due to absorption by gaseous clouds present in front of the background light sources. We point out that this dataset presents a higher level of apparent stochasticity than the previous example and is more difficult to interpret. Here, after reordering the data we also smooth it in the $y$ direction, using a running median filter. this step compensates for the noise, and makes weaker trends more easily apparent. We can observe dark regions corresponding to the absorption of light. The algorithm reveals the existence of two distinct populations of absorbers at the top and bottom of the ordered dataset. This is most obvious when examining the data near $\lambda=1,500\,{\rm \AA}$. These systems are known to exhibit different physical properties (e.g., \citealt{gibson09} and references therein). This example illustrates how the algorithm can naturally perform a clustering task (and define sequences within each cluster), even in the presence of a substantial amount of noise.

\textbf{Images from the natural world.} In the case of simple, highly symmetric physical objects, it is sometimes possible to use a (physical) model and describe each object using a set of parameters. In such a case, it might be possible to identify an underlying trend in the data by looking at a trend in the best fit parameters. When complexity increases, quantitative models based on a small number of parameters are no longer available. The search for trends requires a data-driven approach. To illustrate the performance of the algorithm in a higher-complexity regime where physical modeling of the data is out of reach, we use a set of images from the natural world (natural images). Due to the great variety of shapes and textures, the structural information of such images is expected to be distributed over a wide range of scales and cannot be described by a generic model. To stay in the simpler regime of one-dimensional objects, we randomly shuffle the rows of images and attempt to recover the original ordering. One example is shown in Figure~\ref{f:sequences_physical_datasets}. The left panel illustrates the complexity of each object and the lack of a simple model to characterize them. For the two examples shown, the algorithm is able to recover the original input. In the Appendix we present more examples using natural images and, for comparison, show the corresponding outputs for t-SNE and UMAP. We point out that, for representational purposes, all re-ordered images are flipped so that they match the expected orientation, when necessary. The algorithm has obviously no information regarding the correct orientation of the output. 

\textbf{Displaying the spatial variation of vectors.} Displaying spatial information, for example on a map, is often done through the use of a colorbar, i.e. a sequence. This allows one to display a scalar value as a function of position. If the dynamic range is large, then techniques such as histogram equalization can be used to map the collection of values onto the appropriate range provided by the colorbar. If the quantity to visualize is not a scalar but a vector or a distribution, there is no systematic way to define a mapping onto a colorbar. However, by attempting to order these objects into a sequence, our algorithm provides us with a generic way to meaningfully assign a color to each vector or distribution. The sequencer does to vectors what histogram equalization does to scalars. 

To illustrate how the algorithm can be used to display a set of distributions on a map, we apply it to a dataset from structural seismology. Using the surface wave phase velocity at each of about $14,000$ locations across the contiguous United States \citep{olugboji}, we extract the Love wave dispersion curves specifying velocities at a set of 11 periods, which are primarily sensitive to crustal structure. To homogenize the data, we convert the velocity at each period to its standard score. The ordered-index obtained by the Sequencer can be color-coded. Using the same color bar, we can create a map which naturally shows the geographic patterns in crustal structure, shown in Figure~\ref{f:US_map}.

Prominent regions of thick sediment, such as the Williston and Denver basins and the Mississippi Embayment, are traced out at one end of the sequence. Regions with very contrasting crustal structure, such as the Sierra Nevada, High Rockies, and the Snake River Plain, are traced out at the other end of the sequence. We can also point out the correspondence between the geographic patterns revealed by the Sequencer and physiographic provinces identified at the surface \citep{physio}, shown by black lines in Figure~\ref{f:US_map}. This example illustrates how the proposed algorithm can be used for mapping properties encoded as distributions rather than scalar values.

\section{Discussion}\label{s:discussion}

The elongation of the minimum spanning tree of the distance matrix graph characterizing a dataset can be used to identify which aspects of the data, for example which metric and scales, carry the signatures of an underlying simple trend. 

Applying the Sequencer algorithm to real data has already led to discoveries. In astrophysics, it revealed a sequence in the spectroscopic properties of Active Galactic Nuclei, which led to a novel way to infer the mass of black holes \citep{baron19}. In geology, it led to the detection of seismic waves scattered by previously unrecognized 3D structures near the core-mantle boundary \citep{kim20}. In each case, the data had been publicly available for years but these trends had not been noticed. Once a trend has been identified by the Sequencer, knowing which observational signatures carry relevant information, it is possible to recover the sequence without the aid of the algorithm.

As the use of the elongation of minimum spanning trees allows one to identify a point of view leading to a simple characterization of the data, it can be used more generically to optimize the parameters of any dimensionality reduction technique (e.g. t-SNE or UMAP, see the Appendix for additional details) in order to create a projection of the data that can reveal the meaningful variation.

Ordering a collection of objects often allows some meaning to emerge. For the datasets shown above, the ordered indices could be assigned to physical quantities: we detected sequences in temperature, distance, type of system, and for natural images we recovered height or angle. Having reached this point, it is the start of the scientific analysis requiring specific domain knowledge. The ability of the Sequencer algorithm to reveal the leading trend in a dataset by analyzing pixel data (rather than selected features or summary statistics) can therefore enable or greatly accelerate scientific discoveries.

\subsection{Outlier detection}

As mentioned above, if two types of objects are present in a dataset, the algorithm can identify two distinct clusters and will present them sequentially in the final ordering of the data. This will also apply to cases for which a minority of objects, typically labelled as outliers, differ from the rest of the data. In other words, outliers are typically found at one end of the sequence. One can also imagine some objects for which only a small fraction of the pixels differ from the rest of the population. Such objects or set of pixels would also be labelled as outliers. Interestingly, our approach allows us to detect them in a simple manner. Having an ordered sequence in hand, one can meaningfully perform an averaging/smoothing operation along that dimension and reduce the amount of fluctuations not related to the detected sequence. This has the advantage of enabling the detection of weak trends that are not easily visible in the randomly ordered dataset. Then, once a smooth sequence has been defined, one can look at the pixel-level differences between the original sequence and the smooth counterpart. The statistics of these residuals can reveal objects for which a subset of the pixels differ from the expected underlying trend.

\subsection{Limitations}

A number of considerations must be kept in mind when using the Sequencer: first, the ordering operation performed by the algorithm makes use of one definition of simplicity, based on the elongation of a minimum spanning tree (Eq.~\ref{eq:elongation}). While this always provides a well-defined ordering, it might not necessarily lead to the trend expected from model-based considerations, which are often based on a number of assumptions regarding features, scales, metrics, etc. If prior knowledge is available, it is advised to consider limiting the data to the region(s) where a meaningful variation is expected.
Similary, in some cases, rescaling the values of the input data might lead to better results, especially in cases for which the data present a high dynamic range.

The algorithm attempts to make use of information on different scales which, by default, are logarithmically spaced. While this approach is meant to be generic, it is possible to imagine cases for which this hierarchical decomposition might not be optimal. Segmenting the data using different strategies (for example using wavelets such that Fourier frequency windows do not overlap) might lead to better results. The user can modify the default sampling strategy if need be.

The algorithm attempts to find a trend using all fluctuations present in the data. These fluctuations can be due to useful structural information or due to random noise. Without a model, the algorithm cannot distinguish between them. Noise fluctuations are partially reduced when averaging is performed across segments of the data and scales (Eq.~\ref{eq:combine}) but do present a limitation in identifying the underlying trend. If the dataset presents a range of noise levels, it is advised to first apply the algorithm to a subset of the data with a higher signal-to-noise ratio. As described above, obtaining an ordered sequence offers another opportunity to perform an averaging operation along the sequence and further improve the characterization of the underlying trend.

\section{Conclusions}

Exploratory data analysis, i.e. the search for patterns in datasets without the benefit of guidance from theory, is an important part of scientific research. This search is often challenging due to the apparent complexity of the data and, in practice, it tends to be more of an art than a science.

We present an algorithm, the Sequencer, designed to generically find a continuous sequence in a dataset. Using the shape of graphs characterizing similarities between objects, it can identify which aspects of the data carry the signatures of a simple underlying trend. More specifically, given a metric and a scale, it evaluates the degree to which a continuous trend exists based on the elongation of the minimum spanning tree of the corresponding distance matrix. It can then meaningfully combine this information for a collection of metrics and scales to define a final sequence. By extracting only graph-based geometric information, this method allows us to generically identify aspects of the data that lead to a simple description of the underlying variation.

Using the elongation of a minimum spanning tree as a figure of merit can be used in other contexts. For example, it can be used to optimize the parameters of any dimensionality reduction technique in order to create a projection of the data that reveals meaningful variation. 

To illustrate the power of the algorithm, we have applied it to various datasets with one dimensional objects. It can straightforwardly be applied to objects in two or higher dimensions, but at a higher computational cost. Using scientific datasets and images with randomly-shuffled rows, we have shown that the Sequencer can identify meaningful trends, even if they originate only from parts of the data. In many cases, it is capable of finding sequences that the popular t-SNE and UMAP dimensionality reduction techniques fail to reveal.

Informed by the geometry of graphs, the algorithm provides guidance in extracting simplicity from observations of the complex world. As already demonstrated in astronomy \citep{baron19} and geology \citep{kim20}, the Sequencer can discover unexpected (and sometimes simple) trends in datasets that have already been studied by numerous scientists. This approach to exploratory data analysis, which does not rely on any training nor tuning of parameters, has the potential to enable discoveries in a wide range of scientific domains.

\section*{Materials and Methods}
Code Availability: Implementation details and code are available on GitHub: 
\url{https://github.com/dalya/Sequencer/}. 
An online interface is available at \url{http://sequencer.org}.\\
The algorithm is implemented in {\sc python} and our code relies on the following packages: {\sc numpy} \citep{oliphant06}, {\sc scipy} \citep{scipy01}, {\sc matplotlib} \citep{hunter07}, {\sc networkx} \citep{hagberg08}, and {\sc scikit-learn} \citep{pedregosa11}. The testing and visualization were done with Jupyter notebooks \citep{perez07}.

\acknowledgements{
We thank Vedran Lekic for his guidance and help with the analysis of the seismology data. We thank Manuchehr Taghizadeh-Popp for the implementation of an online platform running the Sequencer algorithm on uploaded datasets. We thank J. Bloom, J. Kaplan, D. Kim, J. X. Prochaska, Y. S. Ting, and S. Zucker for useful comments on the manuscript. This work was supported by the Packard Foundation and the generosity of Eric and Wendy Schmidt by recommendation of the Schmidt Futures program. D. Baron is supported by the Adams Fellowship Program of the Israel Academy of Sciences and Humanities.
}





\onecolumngrid

\section{Algorithm details}

\subsection{list of metrics}
\label{a:distances}

In order to assess the level of similarity between pairs of objects, the algorithm uses a default set of four distance metrics. We briefly describe them here. This set can be expanded by the user if need be. As generically done, we will treat a pair of dataset objects as two probability density distributions $p(i)$ and $q(i)$. An important point to consider is whether the index $i$ encodes a dimension --- and can therefore be shuffled, or the value of a coordinate, such as distance, time, energy, etc. --- in which case the derivative with respect to $i$ carries valuable information. Distance metrics insensitive to shuffling will not take this into account. When the index $i$ encodes the value of a coordinate, for simplicity, we assume that all objects in the set are sampled in the same manner. If not, interpolation/resampling techniques need to be applied prior to using the Sequencer algorithm. The four metrics used by default are:

\begin{itemize}
    \item \textbf{The Euclidean Distance ($L_2$)}, i.e. the familiar distance between two points in an Euclidean space, is the default metric used in many fields. In particular, it is the default metric used by the dimensionality reduction algorithms t-SNE and UMAP. Given two objects that are represented by the vectors $p(i)$ and $q(i)$, $L_2$ is given by:
\begin{equation}\label{eq:l2}
	{\mathrm{L_2}(p,q) = \sqrt{\sum\limits_{i} \Big[ p(i) - q(i) \Big]^2 } }.
\end{equation}
$L_2$ is non-negative, symmetric, and equals to zero only when the two vectors are identical. Furthermore, $L_2$ is insensitive to the relative order of the values in the vector and therefore does not use any (useful) information from the derivatives with respect to index $i$.

\item \textbf{The Kullback-Leibler Divergence (KL-divergence)}, also called the relative entropy, quantifies the degree of surprise involved in seeing one distribution, given another one. It is widely used in Machine Learning to compare the similarity between two objects. For two discrete random variables with PDFs $p(i)$ and $q(i)$, it is given by:
\begin{equation}\label{eq:kl_distance}
	{\mathrm{KL}(p||q) = \sum\limits_{i} p(i)\,\mathrm{log}\frac{p(i)}{q(i)}}
\end{equation}
The KL-divergence is non-negative, and equals to zero only when the two functions are similar in every bin $i$. One can see that it is asymmetric, namely $\mathrm{KL}(p||q) \neq \mathrm{KL}(q||p)$, and it reaches infinity if in at least one bin $i$, $q(i)=0$. We further note that the KL-divergence is not sensitive to the relative order of the bins.

\item \textbf{The Monge-Wasserstein or Earth Mover Distance (EMD):} measures a distance between two distributions from an optimal transport point of view. Intuitively, if the distributions are interpreted as two different ways of piling up a given amount of dirt, the EMD is the minimum work required to turn one pile of dirt into the other, where work is the amount (mass) of dirt moved times the distance by which it moved. 
The EMD is widely used in computer vision, and specifically in content-based image retrieval, as it resembles remarkably well human's visual perception (see e.g., \citealt{rubner00}). 
Interestingly, for one-dimensional distributions, the solution of the optimal transport can be obtained simply by \citep{ramdas15}: 
\begin{equation}
\label{eq:emd_distance}
	{\mathrm{EMD}(p, q) = \int_{-\infty}^{\infty} |P({i}) - Q({i})| d{i}},
\end{equation}
where $P(i)$ and $Q(i)$ are the cumulative distribution functions of the random variables $p(i)$ and $q(i)$.
For higher dimensional distributions, computing the EMD is more computationally demanding. The EMD is non-negative and symmetric. In contrast to $L_2$ and the KL-divergence, the EMD is sensitive to the order and the value of the indices. It can therefore track and quantify translations and/or derivatives with respect to $i$. As a result, the EMD can capture certain aspects of similarities between objects to which $L_2$ and the KL-divergence might not be sensitive.

\item \textbf{The Energy Distance (ED)} provides another measure of statistical distance between two probability distributions. \citet{szekely02} showed that for one-dimensional real-valued random variables, $p(i)$ and $q(i)$, with cumulative distribution functions $P(i)$ and $Q(i)$, it is equivalent to:
\begin{equation}\label{eq:enegy}
	{\mathrm{ED}(p, q) = \Big( 2 \int_{-\infty}^{\infty} |P({i}) - Q({i})| \Big)^2  d{i}  }. 
\end{equation}
The ED is non-negative and symmetric. Similarly to the EMD, it is sensitive to the order and values of the indices.
The differences between the ED and the EMD are analogous to the differences between the $L_1$ and $L_2$ norms. 
\end{itemize}

\subsection{Graph nomenclature}
\label{s:graph_nomenclature}

\begin{figure*}[th]
\begin{center}
\includegraphics[width=.7\textwidth]{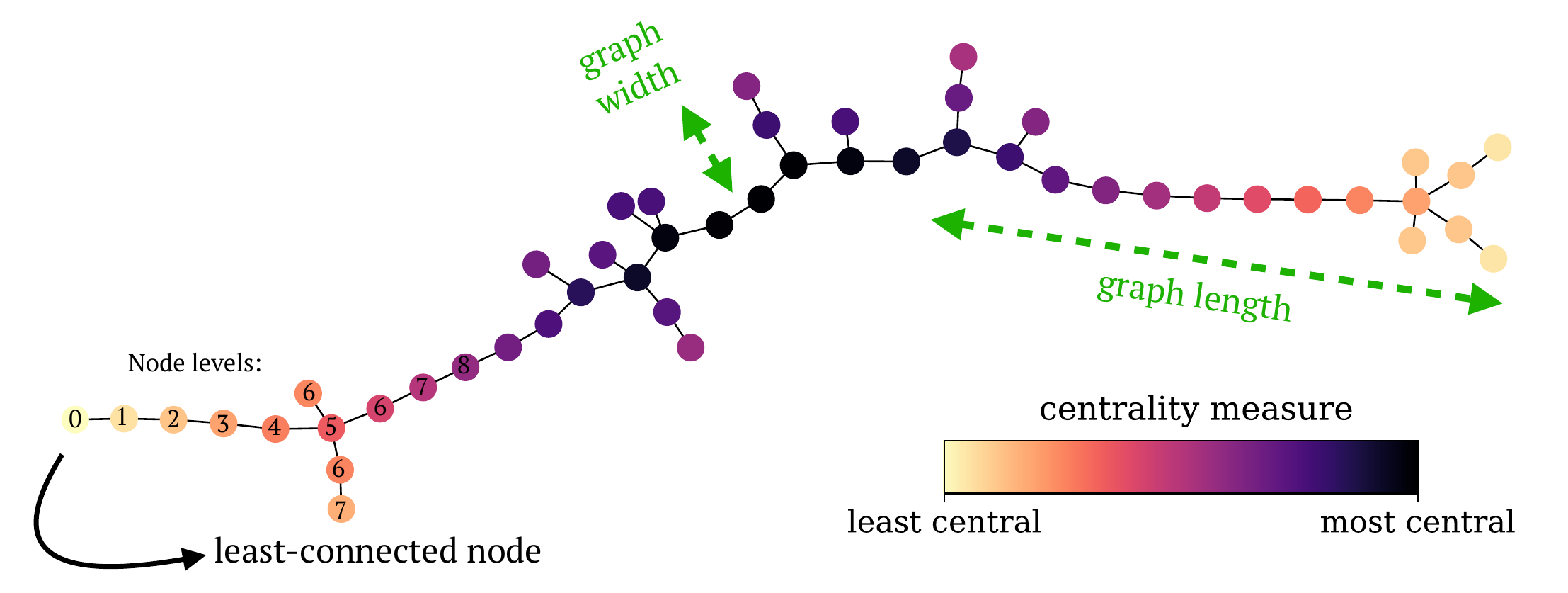}
\caption{{Illustration of various graph minimum spanning tree properties used by the algorithm.}}
\label{f:alg_def_visualization}
\end{center}
\end{figure*}

Here we provide a visual support to summarize the key quantities used by the algorithm and the corresponding terminology introduced in section~\ref{s:algorithm_description}. Figure~\ref{f:alg_def_visualization} shows the example of a minimum spanning tree, where the nodes are color-coded according to their centrality measure. The least connected node of this graph is indicated in the bottom left of the figure. It provides the starting point to traverse the graph using a Bread First Search (BFS) walk, i.e. along the longest branch of the graph and scanning each branch along the way. The nodes are ordered in levels according to their distance (in units of edges) from the starting point (marked with numerical labels in the figure). These levels are used to estimate the average width and height of the tree, which are then used to estimate the elongation of the minimum spanning tree.
The graph length (Eq.~\ref{eq:graph_length}) and width (Eq.~\ref{eq:graph_width}) are also illustrated in the figure.

\subsection{Faster computation for large datasets} 
\label{s:scaling}

The algorithm described in section~\ref{s:algorithm_description} requires distance matrix calculations which scale as $\mathcal{O}(N_{\rm}^2)$. For large datasets this approach can become too computationally demanding. A useful alternative is to first apply the technique to a subset of the data, build the skeleton of a sequence and then populate it with the remaining points. This process leads to a faster computation and enables the analysis of much larger datasets. It can be implemented as follows:
for a dataset with $N_{\rm}$ objects, we first select a subset with $N_s\ll N$ objects for which the distance matrix calculations are computationally feasible. This provides us with a first sequence. 
\begin{itemize}
    \item \textbf{growing sequence:} given a selected sequence, the algorithm selects a fraction $f_A$ of objects distributed uniformly. We refer to them as anchor points. 

    \item\textbf{adding new objects:} to populate the growing sequence with a new object $i$, we first perform a {low-resolution search} where we measure the distance between this object to the $N_A=f_A\,N_s$ anchor points and find the two nearest neighbors. We then perform a {high-resolution search} computing distances between object $i$ and all the nodes from the evolving sequence located between the two nearest anchor points. These distances are populated into the proximity matrix, with a weight that corresponds to the aggregated minimum spanning tree elongation for a given scale (Eq.~\ref{eq:combine_metric}). This proximity matrix is converted into a distance matrix, after which a minimum spanning tree is constructed. We note that, this part of the calculation involves very sparse matrices as only a very small fraction of nodes and edges are considered. This process of populating the growing sequence can be done in parallel for a group of objects. Finally, the updated growing sequence is obtained by a Breadth First Search traversal of the minimum spanning tree. This concludes a single iteration of the population phase.

    \item \textbf{Updated sequence:} after having populated the growing sequence with a group of new objects, we obtain an updated sequence. We can then repeat the process with the remaining points. Here we point out that, at each iteration, the number of \emph{anchor} points, defined to be a fraction $f_A$ of the size of the coarse sequence, grows linearly with the size of the growing sequence. In order to ensure convergence, the number of objects that are populated in every iteration must be smaller than the number of anchor points. 
\end{itemize}
This process avoids a full $\mathcal{O}(N_{\rm}^2)$ calculation and allows one to search for sequences in large datasets in a more efficient manner. This is done at the cost of obtaining only an approximate result for which the accuracy depends on the choice of initial subset size $N_s$ and the fraction $f_A$ of anchor points to use.

\section{Minimum Spanning Tree elongation as a figure of merit}

One of the key ideas presented in this work is that it is possible to quantify the level to which a trend is detectable in a dataset. One can do so by using the elongation of the minimum spanning tree of a distance matrix characterizing this dataset or a collection of distance matrices characterizing various aspects of the dataset. In order to use this method generically and, for example, to compare performances with datasets of different sizes, we can use a normalized elongation parameter:
\begin{equation}
    \eta' = \eta / N
\end{equation}
where $N$ is the number of objects in a dataset. This normalized elongation ranges from $1/N$ for a random graph to $1$ for a perfect sequence. This normalized elongation parameter can be computed for any dataset and for the output of any dimensionality reduction technique. It can thus serve as a figure of merit to quantify the level to which a trend is detectable. It can therefore be used to optimize the parameters of a given algorithm and/or select which technique performs better in order to reveal a continuous trend in a dataset. This procedure could for example be used in combination with the t-SNE and UMAP algorithm to select their internal parameters, metric and possibly the scales of the data leading to the embedding revealing the ``best'' sequence.

\subsection{Comparing the Sequencer to t-SNE and UMAP}
\label{a:comparison_with_t-SNE_and_umap}


In order to compare the performance of the Sequencer to that of t-SNE or UMAP, we need to introduce a procedure to automatically select their parameters leading to the ``best'' sequence but without using the full machinery of the Sequencer. To do so, we proceed as follows:
given a set of hyper-parameters, each of these techniques can be applied to embed a collection of objects or vectors into a one-dimensional space and naturally obtained an ordered list. The question then is whether this list reveals a meaningful trend in the dataset. To address this, it is unfortunately not possible to directly use the elongation parameter in that embedding as, by definition, the corresponding manifold only has one dimension, so $\eta'$ is always one. To meaningfully estimate a corresponding elongation parameter, we need to have access to more information. To do so, we use these embedding techniques to obtain projections in two dimensions. In the corresponding spaces, we can now meaningfully estimate the elongation of these manifolds and use this information as a figure of merit to assess the level to which a sequence is found. Here, we point out that the elongation parameter estimated in this manner can serve as a figure of merit only when the two-dimensional embedding does not strongly depart from a one-dimensional manifold. It is meaningfully defined only when the two-dimensional distribution of objects can be simply represented by a major and minor axis. In cases involving clusters and/or outliers in the two-dimensional projection, the elongation parameter no longer informs on the quality of an expected sequence.

\begin{figure*}[ht]
\begin{center}
\includegraphics[width=.98\textwidth]{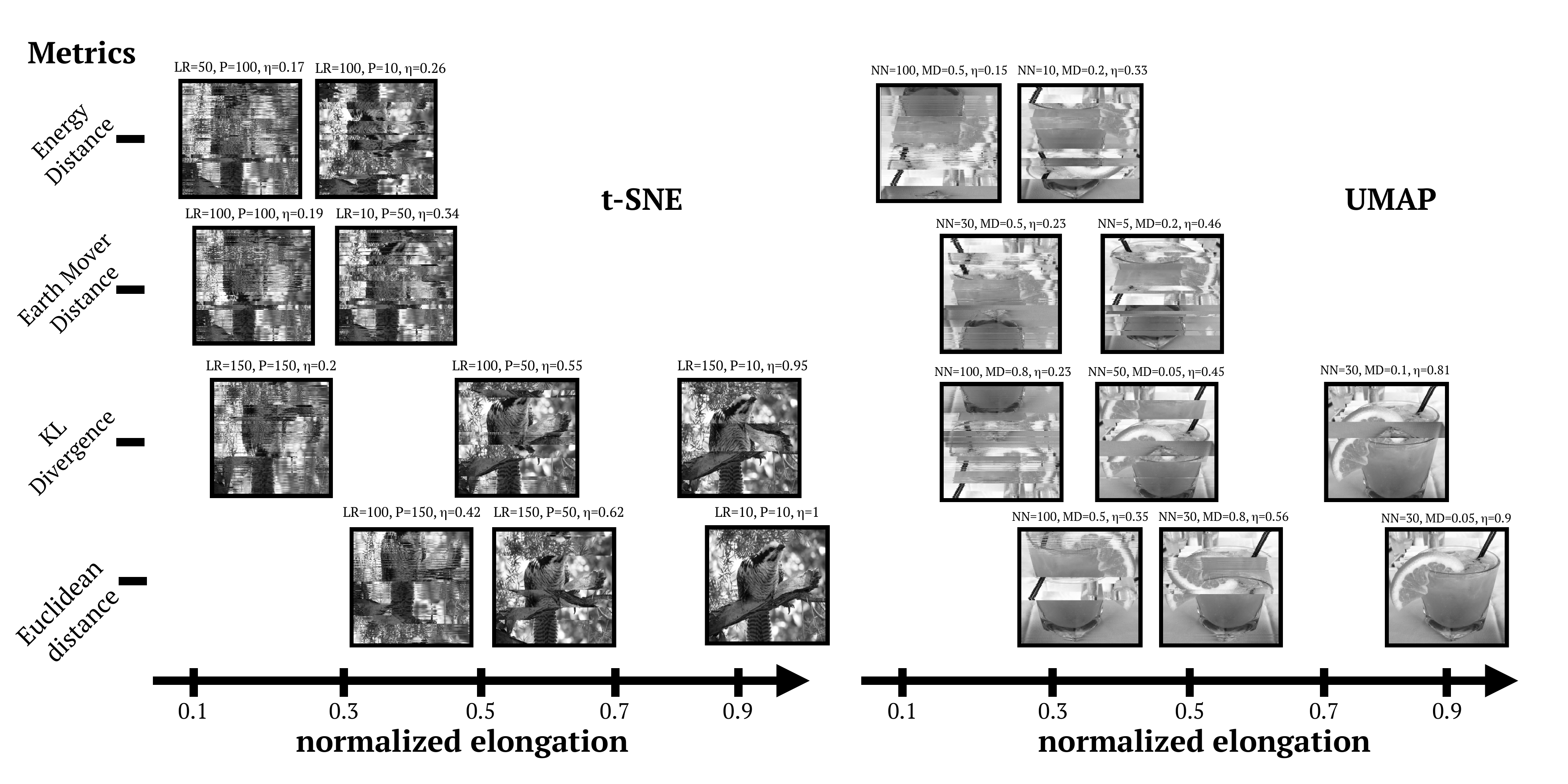}
\caption{
\textbf{Using the elongation parameter as a figure of merit to optimize the hyper-parameters and/or metric of t-SNE and UMAP.} These two embedding techniques are applied to randomly shuffled rows of images. The highest minimum spanning tree elongations naturally selects the correct ordering of the objects within the set.
}
\label{f:t-SNE_hyperpars_L2}
\end{center}
\end{figure*}

As described above, we can then use the elongation parameter as a figure of merit to find the metric and/or hyper-parameters that produce the most elongated sequence or, in other words, the trend presenting the most elongated one-dimensional manifold, i.e. the highest degree of continuity. To illustrate this, we apply t-SNE and UMAP to two images for which the rows have been randomly shuffled. In Figure~\ref{f:t-SNE_hyperpars_L2} we show how varying the hyper-parameters of each technique affects the quality of the resulting embedding and how optimizing for the largest value of the elongation parameter naturally leads to the recovery of the original image. We also point out that optimizing the hyper-parameters of these techniques does not always allow to recover the original image.

\begin{figure*}[ht]
\begin{center}
\includegraphics[width=.8\textwidth]{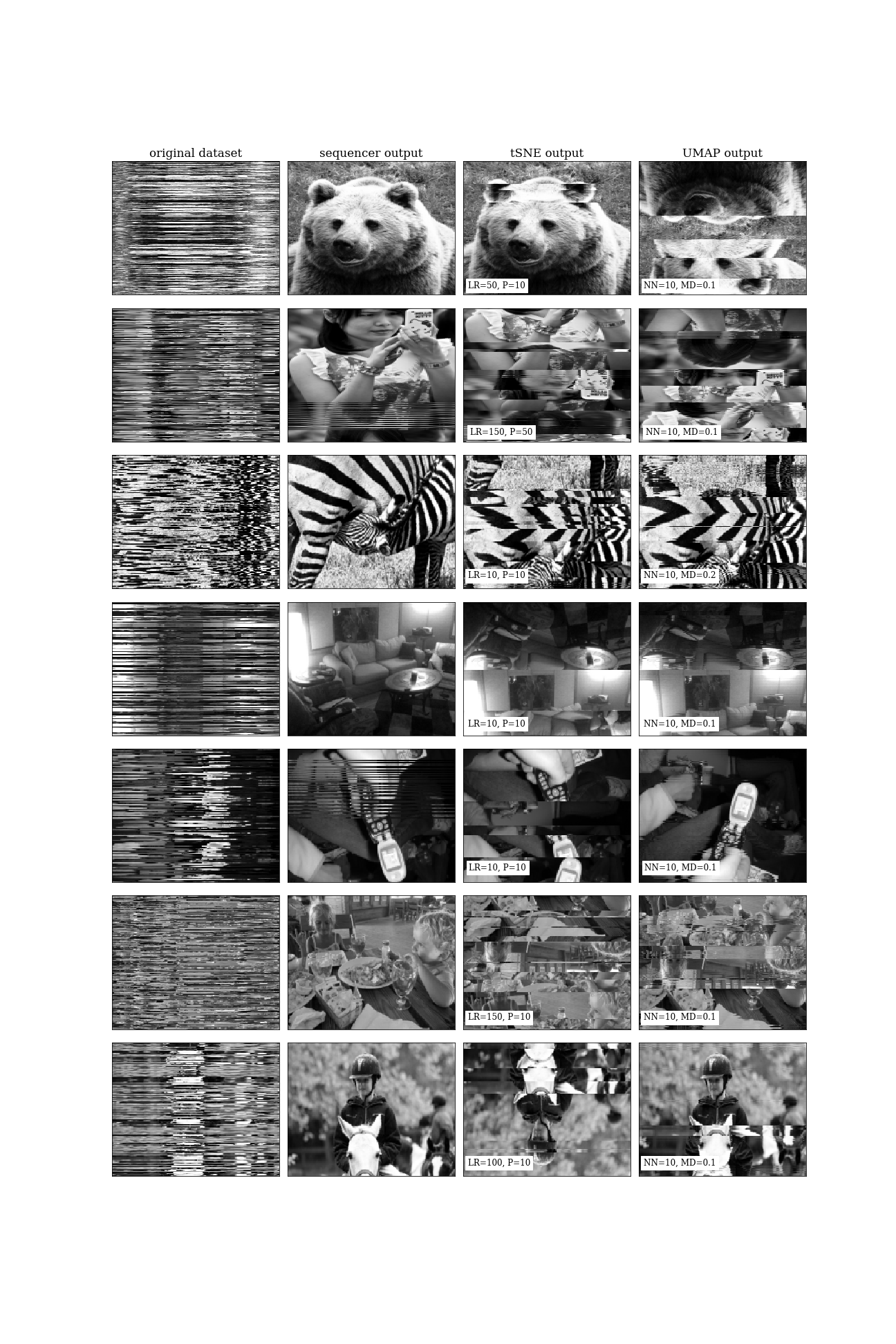}
\caption{\textbf{Comparison between the Sequencer output and the one-dimensional embedding by t-SNE and UMAP for scrambled images}. The first column represents the original scrambled images used as inputs to the three algorithms. The second column shows the reordered image according to the Sequencer's output. The third and forth rows show the reordered image according to the "best" one-dimensional representation by t-SNE and UMAP. For t-SNE and UMAP, at the top of each image we indicate the hyper-parameters that resulted in the highest elongation parameter.}
\label{f:sequences_natural_images_1}
\end{center}
\end{figure*}

Having introduced the normalized elongation parameter to (i) automatically optimize the hyper-parameters and/or metrics of t-SNE/UMAP and (ii) compare the resulting embeddings of different techniques, we now show that the Sequencer algorithm can identify an underlying trend in cases where both t-SNE and UMAP do not. This is similar to the example shown in Section~\ref{f:sequences_natural_images_1} but, this time, we are using real data rather than artificially generated distributions.
To illustrate this point, we select images from the COCO dataset \citep{lin14}. As done previously, we treat the rows as a collection of ordered objects whose order is randomly shuffled. We use them as inputs to the Sequencer, t-SNE and UMAP, and in each case we attempt to recover the original ordering using the procedure described above. For t-SNE, we consider the four distance metrics and the hyper parameters: \texttt{learning\_rate=[10, 50, 100, 150]} and \texttt{perplexity=[10, 50, 70, 100]}. For UMAP, we consider the same distance metrics and the hyper-parameters: \texttt{n\_neighbour=[10, 50, 100, 150]} and \texttt{min\_dist=[0.1, 0.2, 0.5, 0.8]}. We note that this coarse sampling of the hyper parameters of these two techniques appears to be enough for this type of data. A higher resolution optimization 
of the parameters does not provide better results.

The results are shown in Figure~\ref{f:sequences_natural_images_1}. For each input shuffled image, we present the output of the Sequencer, as well as the "best" outputs obtained with t-SNE and UMAP. In all cases, the elongation-based optimization leads to meaningful segments of the original images. Interestingly, in a number of cases, the Sequencer outperforms the best possible outputs obtained with t-SNE and UMAP. As described in section~\ref{s:performance:simulated_dataset}, this is due to the ability of the Sequencer to look for a global trend using multiple metrics and a range of scales.

\bibliography{main-ref.bib}



\end{document}